\documentclass[sn-mathphys,Numbered]{sn-jnl}

\usepackage{graphicx}%
\usepackage{multirow}%
\usepackage{amsmath,amssymb,amsfonts}%
\usepackage{amsthm}%
\usepackage{mathrsfs}%
\usepackage[title]{appendix}%
\usepackage{xcolor}%
\usepackage{textcomp}%
\usepackage{manyfoot}%
\usepackage{booktabs}%
\usepackage{algorithm}%
\usepackage{algorithmicx}%
\usepackage{algpseudocode}%
\usepackage{listings}%

\usepackage{adjustbox}
\usepackage{multicol} 
\usepackage{multirow}
\usepackage{array}
\usepackage{tabularx}

\usepackage{amsmath}
\numberwithin{equation}{section}

\theoremstyle{thmstyleone}%
\theoremstyle{thmstyletwo}%

\theoremstyle{thmstylethree}%

\begin{document}

\title[Article Title]{FaceGemma: Enhancing Image Captioning with Facial Attributes for Portrait Images}








\author[1]{\fnm{Naimul} \sur{Haque}\email{naimul011@gmail.com}}
\author[2]{\fnm{Iffat} \sur{Labiba}\email{iffatlabiba.miu@gmail.com}}
\author[2]{\fnm{Sadia} \sur{Akter}\email{sadiaakterakter178@gmail.com}}
\affil[1]{\orgdiv{Department of CSE}, \orgname{Uttara University}, \orgaddress{\country{Bangladesh}}}
\affil[2]{\orgdiv{Department of CSE}, \orgname{Manarat International University}, \orgaddress{\country{Bangladesh}}}


\abstract{Automated image caption generation is essential for improving the accessibility and understanding of visual content. In this study, we introduce FaceGemma, a model that focuses on accurately describing facial attributes such as emotions, expressions, and features. Using FaceAttdb data, we generated descriptions for 2000 faces with the Llama 3 - 70B model and fine-tuned the PaliGemma model with these descriptions. Based on the attributes and captions supplied in FaceAttDB, we created a new description dataset where each description perfectly depicts the human-annotated attributes, including key features like attractiveness, full lips, big nose, blond hair, brown hair, bushy eyebrows, eyeglasses, male, smile, and youth. This detailed approach ensures that the generated descriptions are closely aligned with the nuanced visual details present in the images. Our FaceGemma model leverages an innovative approach to image captioning by using annotated attributes, human-annotated captions, and prompt engineering to result in high quality facial descriptions. Our method significantly improved caption quality, achieving an average BLEU-1 score of 0.364 and a METEOR score of 0.355. These metrics demonstrate the effectiveness of incorporating facial attributes into image captioning, providing more accurate and descriptive captions for portrait images.}




\maketitle

\section{Introduction}\label{sec1}
Image captioning \cite{imagecaptioning} represents a captivating and multifaceted interdisciplinary challenge, uniting the realms of Computer Vision \cite{computervision} and Natural Language Processing (NLP) \cite{nlp}. Its primary objective is to create meaningful and contextually relevant captions for images by using advanced Deep Learning models \cite{dl}. These models are pivotal in extracting important visual features and understanding the semantic context necessary for generating informative and coherent human-readable captions. With far-reaching applications that include aiding the visually impaired
and enriching human-computer interactions, image captioning stands at the forefront of AI research.

The landscape of automatic image captioning has witnessed remarkable progress, largely attributed to the advent of Deep Neural Networks and the availability of extensive captioning datasets. These networks are renowned for producing fact-based image descriptions. Recent endeavors in this domain have pushed the boundaries, focusing on detecting emotional and relational facets within images and infusing captions with emotive features to craft more engaging and immersive descriptions. In summation, image captioning remains a pivotal enabler of AI technology, narrowing the gap between visual content and human language to facilitate seamless human-machine interactions.

In the realm of modern Image Captioning, an encoder-decoder paradigm 
\cite{en-dec3,en-dec4}
is commonly adopted. This top-down approach begins with a Convolutional Neural Network (CNN) \cite{cnn} model performing image content encoding, followed by a Long Short-Term Memory (LSTM) \cite{lstm} model which is responsible for generating the image caption through decoding. Based on a thorough review of the current state-of-the-art, we have decided to develop our Facial Attribute Image Captioning Model, FaceGemma, following this paradigm.

Several image captioning systems have addressed facial expressions and emotions, but facial attribute captioning goes further by capturing and describing specific physical characteristics beyond emotions. To our knowledge, despite the presence of some facial attribute recognition models, there is no image captioning system that can generate captions based on the attributes of a subject's face from an image. This gap in the literature has motivated our work on building a Facial Attribute Image Captioning Model, named FaceGemma.

In this paper, we also introduce the dataset FaceAttDB \footnote{\url{https://www.kaggle.com/datasets/naimul314/faceattdb}} carefully curated and generated by our research team. This dataset contains comprehensive descriptions of 40 facial attributes such as attractiveness, full lips, big nose, blond hair, brown hair, bushy eyebrows, eyeglasses, male, smile, and youth, among others. We further used these captions and attributes to generate new descriptions of the images using Llama3 70B parameterized model and prompt engineering. It serves as a foundation for our proposed model, FaceGemma \footnote{\url{https://huggingface.co/naimul011/FaceGemma}}, which leverages the synergy of Computer Vision and Natural Language Processing to automatically recognize and describe this diverse range of facial attributes within images. Our contributions can be summarized as follows:

\begin{enumerate}
    \item The FaceGemma model, takes a pioneering approach by generating portrait image descriptions that seamlessly incorporate a wide range of facial attributes. To the best of our knowledge, this represents the first study to apply facial attribute analysis within the realm of image captioning.
    
    \item We evaluate our FaceGemma model through a set of experiments utilizing well-established evaluation metrics, namely BLEU and METEOR. Furthermore, we systematically compare the impact of several image feature extraction techniques on our model's performance.
    
    \item The 'FaceAttDB' image caption dataset, comprising portrait images sourced from the CelebA dataset \cite{faceattributes}, accompanied by captions. We have made this dataset publicly available\footnote{\url{https://zenodo.org/record/8144361}} to support future research in this field.
\end{enumerate}

The remainder of the paper is structured as follows: Section \ref{sec2} covers related works, Section \ref{sec3} introduces our dataset, Section \ref{sec4} details the facial description generation which is used for the final method, Section \ref{sec5} presents our proposed methodology, Section \ref{sec6} presents our results, and finally, Section \ref{sec8} concludes the paper.

\section{Related Works}\label{sec2}

In the intersection of Computer Vision and Natural Language Processing, the field of image captioning for facial images offers many opportunities. It can enhance human-computer interaction, help understand emotional expressions, and support assistive technologies. Although research on captioning facial attributes is still new, it builds on established areas like general image captioning, facial expression analysis, and face recognition.

In face recognition, Parkhi et al.\cite{deepface} (2015) developed a deep neural network that learns discriminative features from facial images, achieving high accuracy despite challenges like lighting and pose variations, but requiring extensive training datasets and computational resources. Nezami et al.\cite{sentiattend} (2018) introduced SENTI-ATTEND, integrating sentiment analysis and attention mechanisms into image captioning to produce emotionally meaningful and contextually relevant captions. Fuhai Chen et al.\cite{groupcap} (2018) proposed "GroupCap," a framework for group-based image captioning that uses structured relevance and diversity constraints to create coherent and diverse captions for groups of images, improving the quality and richness of image descriptions.

In 2019, Nezami et al. introduced Face-Cap\cite{facecap}, a method that combined facial expression analysis with caption generation, achieving better performance than traditional methods. Later in 2019, Hong et al. introduced the Facial Expression Sentence (FES) Generation\cite{facetells}, which associated facial action units with natural language descriptions to capture facial expressions in image captions. In 2020, FACE-CAP and FACE-ATTEND\cite{faceattend} were presented, integrating facial expression features into caption generation to create emotionally resonant captions.

Facial Recognition for Identity Cards \cite{electronicidentity} Usgan et al. made significant contributions in 2020 by improving facial recognition for electronic identity cards, resulting in enhanced identification accuracy. Moreover, Tran et al Entity-Aware News Image Captioning \cite{transformtell}, (2020) demonstrated an innovative approach by integrating named entity recognition with transformer-based caption generation, leading to more informative image descriptions.

In 2021, BORNON\cite{bornon} improved Bengali image captioning. In 2022, Priya S. enriched captions with emotional cues Emotion-Based Caption Generation \cite{cspdensenet}, while Bisikalo explored emotional attitudes \cite{explainingemotion}. Object-centric unsupervised Captioning\cite{unsupervised}, comprehensive facial representation, and fair attribute classification advanced the field.

A 2022 study integrated visual and linguistic cues for comprehensive facial representation, aiding facial attribute prediction and emotion recognition named by General Facial Representation Learning \cite{general} in 2022. Park et al. proposed Fair Contrastive Learning for Facial Attribute Classification \cite{attributeclassify} addressed biased learning in facial attribute classification, achieving fairer results. In 2023, EDUVI \cite{eduvi} combined Visual Question Answering and Image Captioning to enhance primary-level education through interactive and informative mechanisms.

While these studies contribute significantly to image captioning, a notable gap exists in the exploration of facial attributes in portrait image captioning, presenting an opportunity for future research, such as the proposed FaceGemma model.

\section{Dataset}\label{sec3}

Our dataset comprises 2,000 curated portrait images, sourced from the CelebA dataset \cite{faceattributes}, which boasts over 200,000 celebrity images with 10,177 unique identities and various facial attributes. CelebA supports numerous computer vision tasks. Each image in our dataset is paired with five captions.

\begin{figure}[h!]
\begin{center}
  \includegraphics[width=\linewidth]{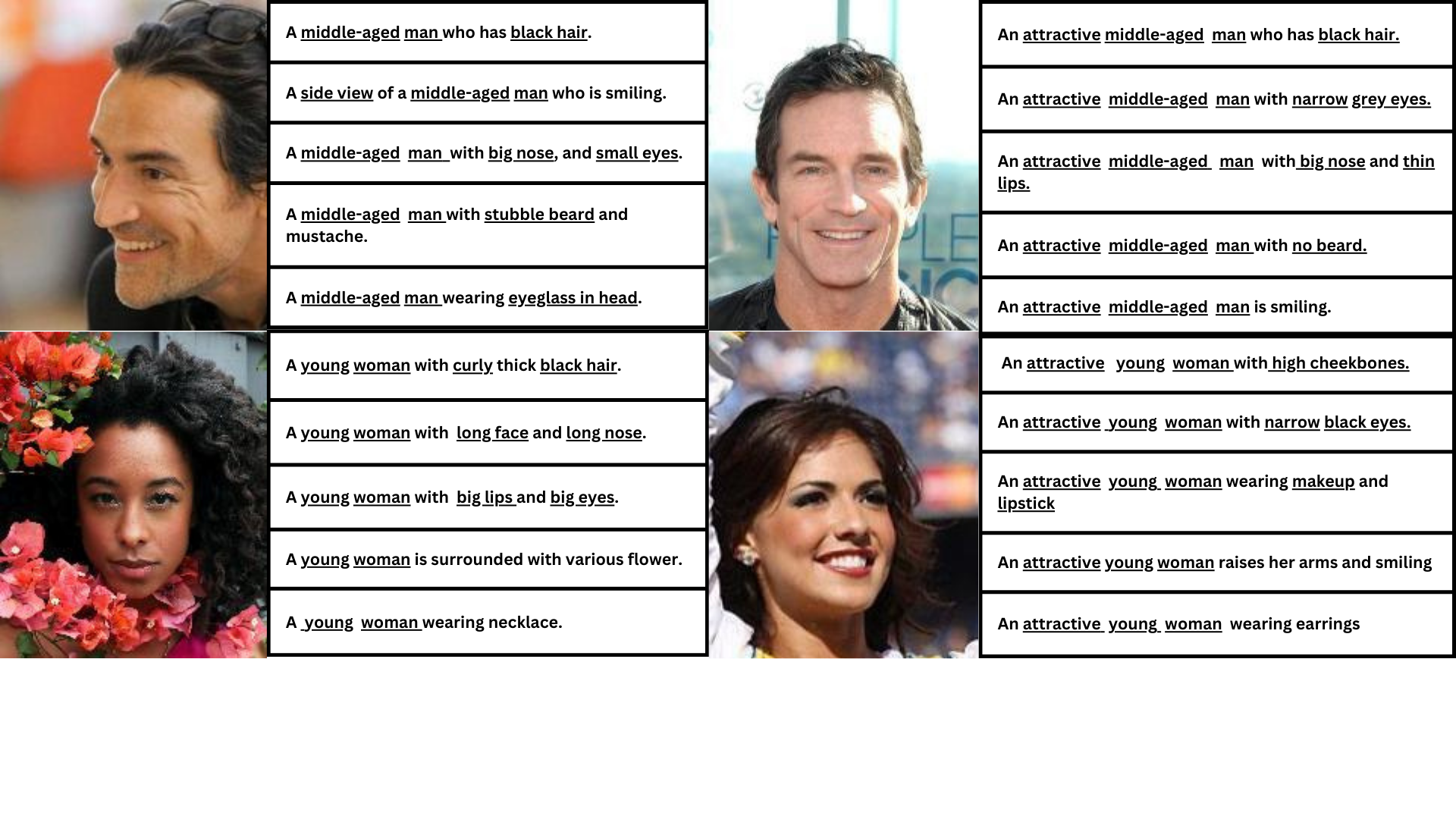}

  \caption{A partial example of our FaceAttDB dataset \cite{faceattdb} where we have highlighted the attributes in each caption.}

  \label{Dataset1}
   \end{center}
 \end{figure}

The captions in this dataset are generated based on the 40 attribute annotations available in the CelebA dataset, which covers various aspects such as age, gender, expression, hair color, eye color, nose shape, and skin complexion. These attributes are used to create captions that accurately describe the visual characteristics of the portrait images. We have included a figure displaying examples of our generated captions, in which we emphasize and highlight the various facial attributes Figure \ref{Dataset1}. This visual representation serves to illustrate the accuracy and effectiveness of our image captioning model in capturing and describing facial features, contributing to a better understanding of its performance


It is important to note that while the FaceAttDB dataset currently comprises 2,000
images, each with five captions, the size of the dataset may be expanded in future work to
enhance the diversity and robustness of the models trained on it which ensures its continued usefulness for advancing research in facial attribute captioning.

\section{Facial Description Engineering}\label{sec4}
The captions that are annotated by humans do not contain all or most of the attributes from our attribute vector list. Hence, we had to engineer prompts that effectively generate facial descriptions using human captions, attributes, and prompt engineering. The engineered prompts are used to generate facial descriptions using the Llama3 large language model.
\subsection{Llama3}
Meta's Llama3 \cite{llama3} is an advanced open-source large language model (LLM) available in two sizes: 8 billion and 70 billion parameters. It excels in comprehending and generating intricate text, making it highly suitable for tasks such as image captioning. Although there isn't an official research paper published yet, benchmarks indicate that Llama3 surpasses most other open-source models of similar scale. This superiority can be attributed to its training on an extensive dataset, which is seven times larger than its predecessor, and advancements in training methodologies. Llama3 offers superior performance for text generation tasks compared to currently available open-source alternatives.

In this study, we employed the Llama 3 70B model which is 3 generation iteration of Meta's Llama language model \cite{touvron2023llama} to enhance the generation of facial attribute descriptions within the FaceAttDB dataset. This model was instrumental in generating detailed descriptions for 2000 faces, leveraging advanced prompt engineering and attribute annotations. The Llama 3 70B model facilitated the creation of a new dataset where each description meticulously portrays human-annotated attributes such as attractiveness, full lips, big nose, blond hair, brown hair, bushy eyebrows, eyeglasses, male, smile, and youth.

\begin{figure}[h!]
\begin{center}
  \includegraphics[width=\linewidth]{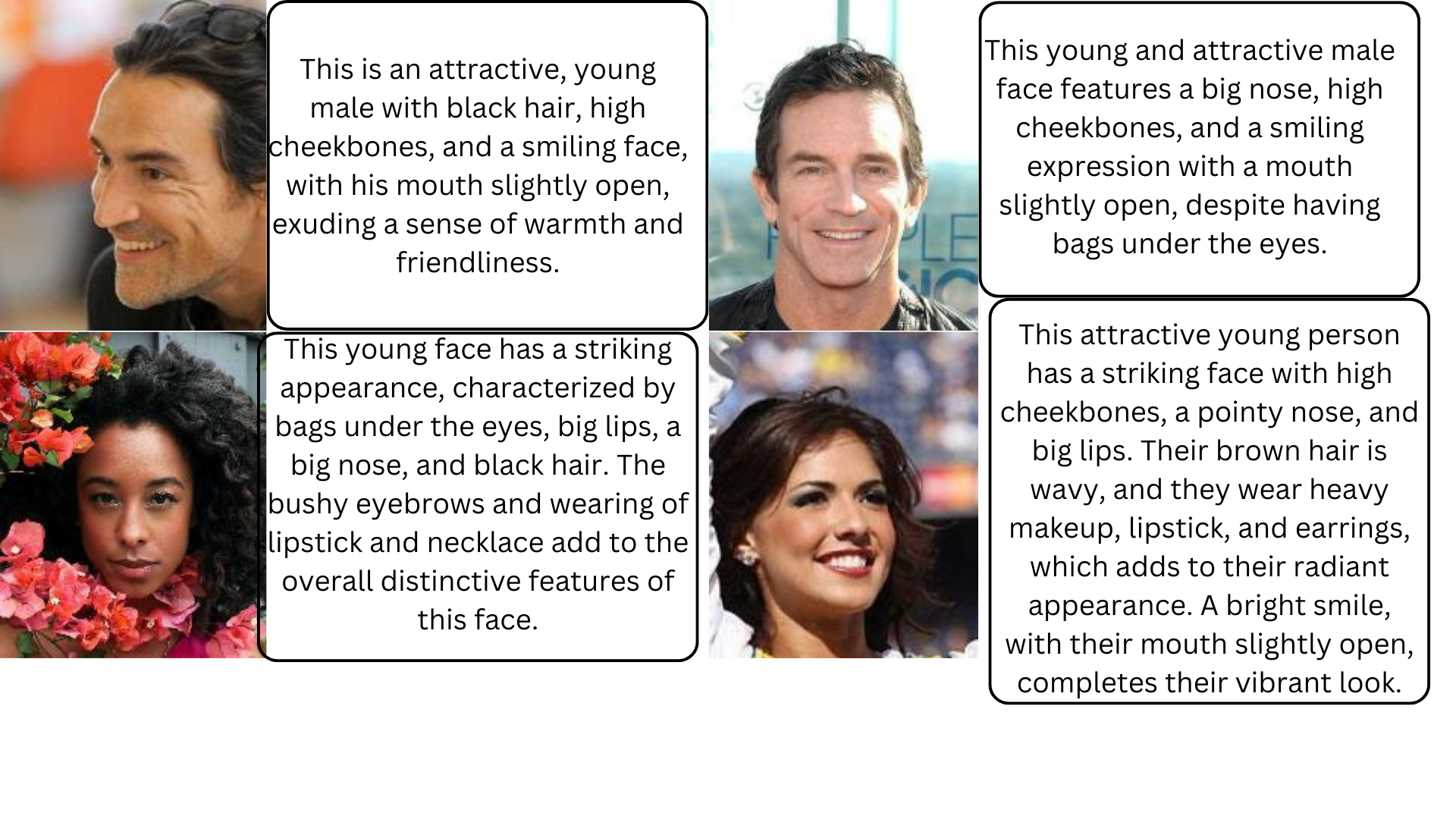}

  \caption{A partial example of generated facial descriptions by our model FaceGemma which has covered the various range of facial attributes.}

  \label{descriptions}
   \end{center}
 \end{figure}

By leveraging the Llama 3 70B model, we ensured that the generated descriptions accurately capture nuanced visual details present in portrait images. These descriptions served as crucial training data for fine-tuning our PaliGemma model, a 2 billion parameter model designed to excel in facial attribute captioning. Figure \ref{descriptions} shows a few of the samples of the generated facial description.

\section{Proposed methodology}\label{sec5}

In this section, we outline the key processes of our FaceGemma model which is our fine-tuned PaliGemma model. We start by explaining the interworkings of PaliGemma and then how this multimodal model inference on the image using prompts.

\subsection{PaliGemma}
PaliGemma \cite{paligemma} is a vision-language model that processes both image and text inputs, inspired by PaLI-3 \cite{chen2023pali3visionlanguagemodels}  and incorporating SigLIP \cite{zhai2023sigmoidlosslanguageimage} and Gemma \cite{gemmateam2024gemmaopenmodelsbased} components. It answers questions about images, generates captions for visual media, detects objects, and reads embedded text. It offers general-purpose pre-trained models adaptable through fine-tuning for diverse applications, alongside research-oriented models specialized for specific datasets. Fine-tuning optimizes PaliGemma for specific tasks, enhancing accuracy and relevance in both general and research contexts, making it versatile for detailed image analysis and multimodal understanding.

\subsection{Prompt Engineering}
To generate new descriptions \( D_{*}(i) \) from the dataset \( D(A_{i}, C_{i}) \), which contains attributes \( A_{i} \) and captions \( C_{i} \), we employ prompt engineering \( PE(x_{i}, \text{prompt}) \). Here, \( x_{i} \) represents each sample caption-attribute pair from \( D \) which can be represented as \( x_{i} = (C_{i}, A_{i})\). The process involves applying prompt engineering to each \( x_{i} \), using a specific prompt, to produce descriptions that are tailored to the attributes and captions provided in the dataset \( D(A, C) \).

\subsection{Description Generation}
The equation shows mathematically how we generate facial descriptions for the images, using prompt engineering and language model $f(.)$ function:

\[ D_{*}(i) = f(\text{PE}(x_{i}, \text{prompt})) \]

This formulation ensures that the generated descriptions \( D_{*}(i) \) reflect the nuanced details of the attributes and captions present in the original dataset, thereby enhancing the accuracy and relevance of the generated text outputs using Llama3's 70B $f(.)$ parameterized model.

Our engineered prompt instructs the Llama3 model to expertly analyze attributes and captions, and craft a coherent paragraph description of the face. Key guidelines include combining positive attributes with "and" for clarity, excluding attributes with a value of 0.0, in the attribute vector, to focus on significant features, and emphasizing attributes labeled at 1.0. The goal is to ensure the descriptions are fluent and natural, resembling human-generated text.
    
\subsection{Method}\label{sub5}
\begin{figure}[h!]
\begin{center}
  \includegraphics[width=\linewidth,]{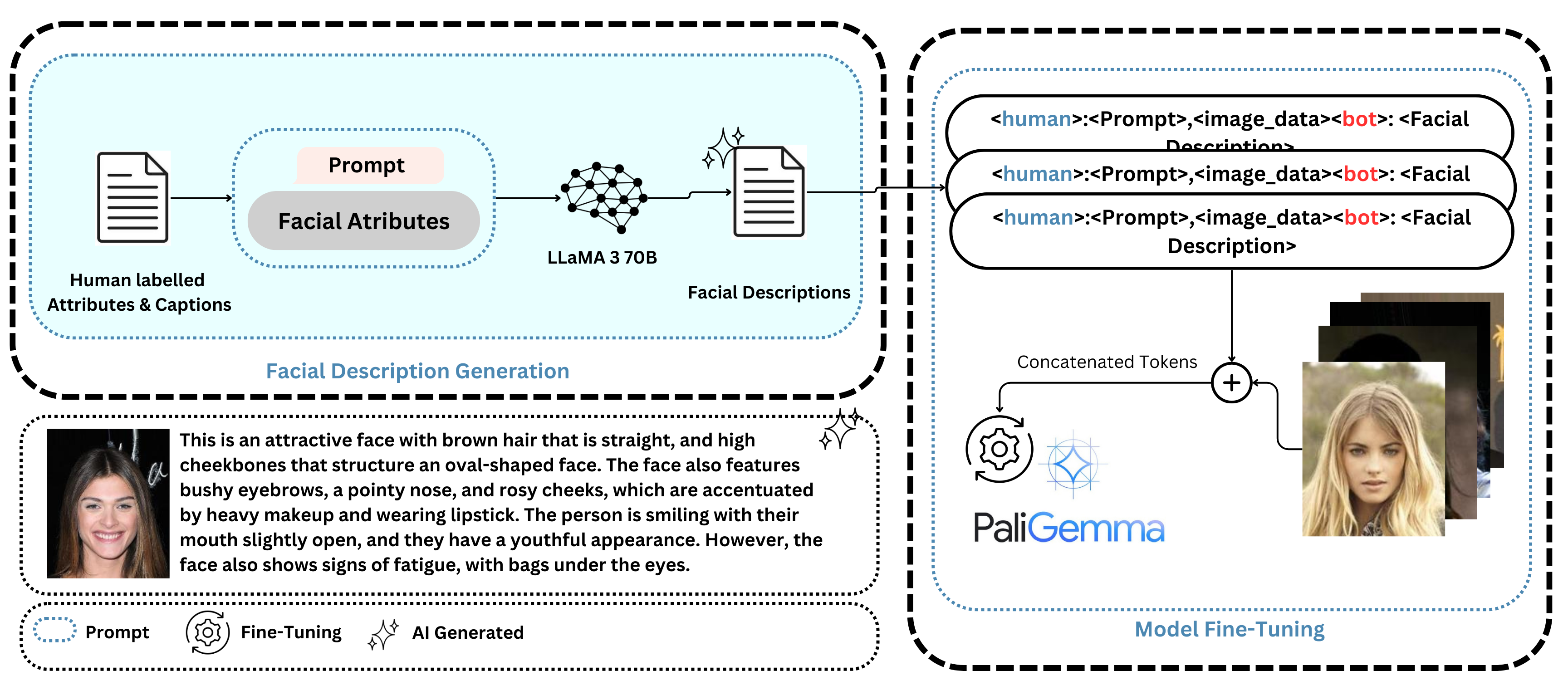}
  
  \end{center}
  \caption{Methodology of FaceGemma. The Llama3 70B model generates detailed descriptions for 2000 faces in the FaceAttDB dataset, focusing on specific facial attributes. These descriptions form a new dataset, which, along with the corresponding images, is used to fine-tune the PaliGemma model. The fine-tuning process pairs each image with its corresponding description, enabling PaliGemma to learn to align its outputs with human-written descriptions. The final model, FaceGemma, is then evaluated for its effectiveness in accurately describing facial features in unseen images.}
  \label{fig:method}
 \end{figure}

Our overall method is shown in figure \ref{fig:method} where the first step of facial description generation has been shown which is also explained in the previous subsection section.

Next, the generated description dataset $D_{*}(i)$ and corresponding image dataset $I_{i}$ are used for fine-tuning the PaliGemma model to form FaceGemma. This fine-tuning takes center stage in transforming PaliGemma into FaceGemma. Each image, $I_{i}$, and description $D_{*}(i)$ is paired with its corresponding description, and PaliGemma learns to bridge the gap between its own generated descriptions and the human-written ones in $D^*(i)$. This iterative process involves minimizing a loss function that measures this difference. Finally, FaceGemma's newfound expertise is evaluated on a separate dataset to ensure it can describe facial features effectively in unseen images. Through this focused training, FaceGemma leverages PaliGemma's foundation and the rich details in $D^*(i)$ to become a master of facial attribute description.

\subsection{Inference}\label{subsec2}

\begin{figure}[h!]
\begin{center}
  \includegraphics[width=\linewidth,]{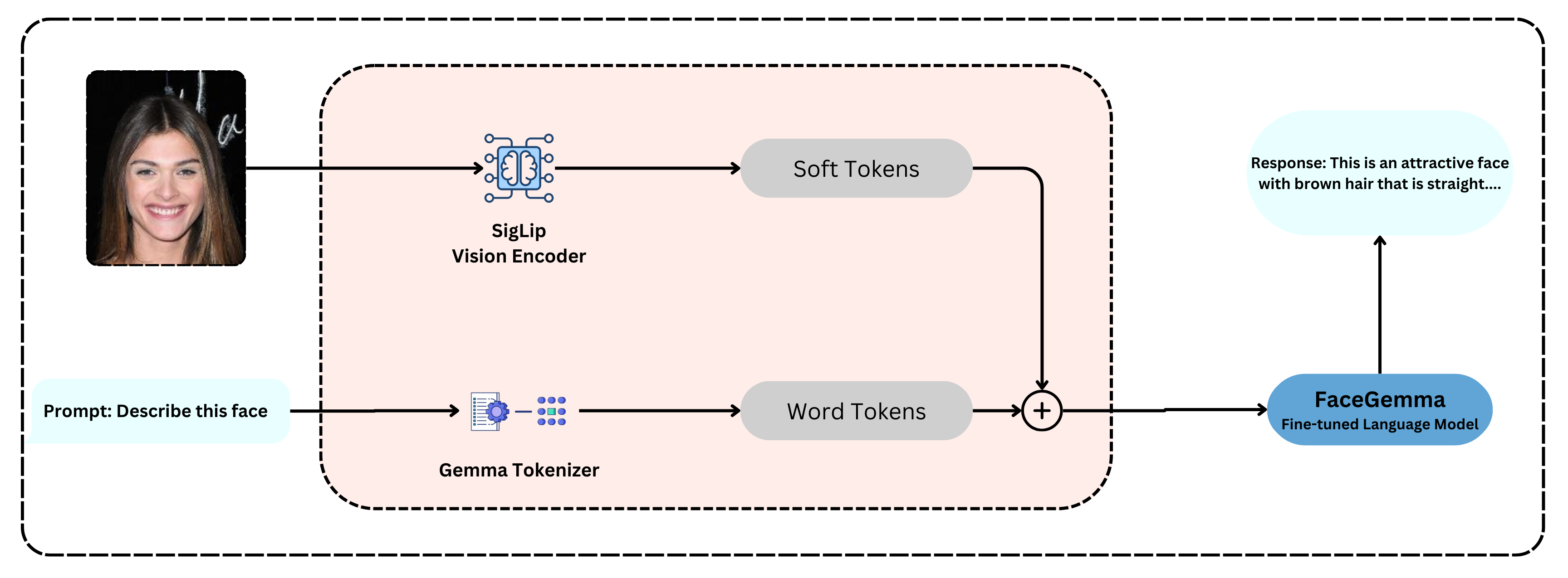}
  
  \end{center}
  \caption{The inference process of the fine-tuned FaceGemma model. The testing image is converted into soft tokens using the SigLip visual feature extractor. Simultaneously, the prompt is tokenized into word tokens by Gemma. These image features and word tokens are concatenated and passed to the fine-tuned Gemma model to generate a descriptive response for the portrait image.}
  \label{fig:facegemma}
 \end{figure}

The inference process of the fine-tuned FaceGemma has been depicted in the figure \ref{fig:facegemma}. The testing image is first converted into soft tokens using the visual feature extractor model SigLip. Simultaneously the prompt is supplied which is tokenized by Gemma's tokenization method into word tokens. These soft tokens or image features are concatenated with the word tokens and passed to the fine-tuned Gemma model to get a descriptive response of the supplied portrait image.

\subsection{Training FaceGemma Model}\label{subsec3}
Before training, the batch size, number of training examples, and learning rate hyperparameters are tuned to improve the model's performance. In the experiment you described, a batch size of 4, 512 training examples, and a learning rate of 0.003 were used.

\begin{figure}[h!]
\begin{center}
  \includegraphics[width=8cm, height=5cm]{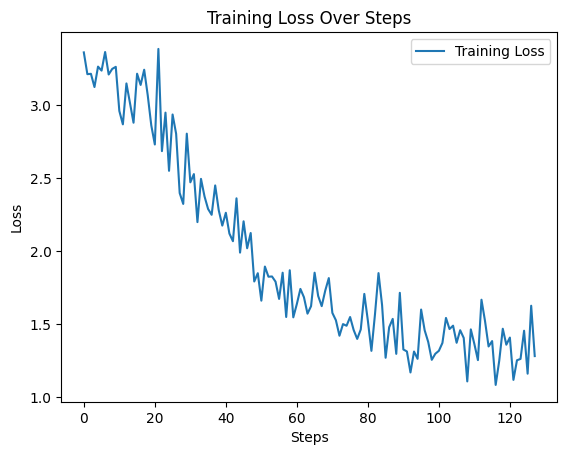}
  
  \end{center}
  \caption{
The figure presents a graph depicting the training loss over time for the PaliGemma model, which is used to generate descriptions of portrait images. The x-axis denotes the number of training steps, while the y-axis indicates the training loss. This loss is determined by comparing the model's generated descriptions with the actual ground truth descriptions. A decrease in training loss signifies improved model performance.}
  \label{fig:training}
 \end{figure}

Figure \ref{fig:training} shows a graph of training loss over steps for training PaliGemma, a model for generating descriptions of portrait images. The x-axis represents the number of training steps, while the y-axis represents the training loss. It is calculated by comparing the model's generated descriptions to the ground truth descriptions. A lower training loss indicates that the model is performing better.

The graph shows that the training loss decreases over time, which means that the model is gradually learning to generate better descriptions of portrait images.

\section{Result}\label{sec6}

We evaluated our model FaceGemma's predicted captions using BLEU and METEOR scores.


\textbf{BLEU}
(Bilingual Evaluation Understudy) \cite{bleu} is a metric for evaluating machine-generated translations by measuring the similarity between generated and reference text using n-gram overlap. The BLEU-n score is calculated as in Eq. \ref{eq:8}. 

\begin{equation}
    \label{eq:8} \text{BLEU} = \text{BP} \times \exp\left(\frac{1}{N} \sum_{n=1}^{N} \log p_n\right)
\end{equation}

$\text{BP}$ is the Brevity Penalty, $N$ is the maximum n-gram order (typically 4), and $p_n$ is the precision of n-grams. $p_n$ and BP are calculated as in Eq. \ref{eq:9} and Eq. \ref{eq:10}. 

\begin{equation}
    \label{eq:9} p_n = \frac{\text{Matching n-grams}}{\text{Total n-grams}}
\end{equation}

\begin{equation}
    \label{eq:10} BP = \begin{cases} 1 & \text{if } c > r \\ e^{(1 - \frac{r}{c})} & \text{if } c \leq r \end{cases}
\end{equation}

BLEU ranges from 0 to 1, with higher scores indicating better translations. Different BLEU variants (BLEU-1, BLEU-2, BLEU-3, BLEU-4) consider different n-gram orders.

\textbf{METEOR}
(Metric for Evaluation of Translation with Explicit Ordering) \cite{meteor} is another metric for translation quality. It considers precision, recall, and $\alpha$ to balance them: 

\begin{equation}
    \label{ea:11} \text{METEOR} = \frac{(1 - \alpha) \times \text{precision} \times \text{recall}}{(1 - \alpha) \times \text{precision} + \alpha \times \text{recall}}
\end{equation}

Precision and recall involve n-gram matching and $\alpha$ typically equals 0.9. METEOR provides a comprehensive evaluation of translation quality, including word order and vocabulary differences.

\begin{table}[htbp]
    \centering
    \caption{Comparison of the scores of our final model FaceGemma with other techniques}
    \label{table:epoch_100}
    \begin{tabular}{@{}ccccccc@{}}
        \toprule
        \textbf{Model} & \textbf{BLEU-1} & \textbf{BLEU-2} & \textbf{BLEU-3} & \textbf{BLEU-4} & \textbf{METEOR} \\
        \midrule
        VGGFace & 0.3357 & 0.0849 & 0.0252 & 0.0064 & 0.2670 \\
VGGFace + Att & 0.3737 & 0.0881 & 0.0222 & 0.0034 & 0.2776 \\
ResNet50 & 0.3408 & 0.0725 & 0.0181 & 0.0034 & 0.2674 \\
ResNet50 + Att & \textbf{0.3900} & 0.0849 & 0.0219 & 0.0060 & 0.3048 \\
InceptionV3 & 0.3050 & 0.0601 & 0.0144 & 0.0015 & 0.2083 \\
InceptionV3 + Att & 0.0000 & 0.0000 & 0.0000 & 0.0000 & 0.0000 \\
\textbf{FaceGemma} (our) & 0.3641 & \textbf{0.2349} & \textbf{0.1514} & \textbf{0.0988} & \textbf{0.3550} \\
        
        \bottomrule
    \end{tabular}
\end{table}

Our experiments involved comparing FaceGemma with several baseline models, namely VGGFace, ResNet50, and InceptionV3, each augmented with LSTM language models and facial attribute vectors. Table \ref{table:epoch_100} presents a comparison of BLEU and METEOR scores between FaceGemma and the baseline models. FaceGemma achieved notable improvements, particularly in BLEU-2 (0.2349), BLEU-3 (0.1514), BLEU-4 (0.0988), and METEOR (0.3550) metrics, surpassing all other techniques evaluated.



\begin{figure}[h!]
\begin{center}
  \includegraphics[width=\linewidth]{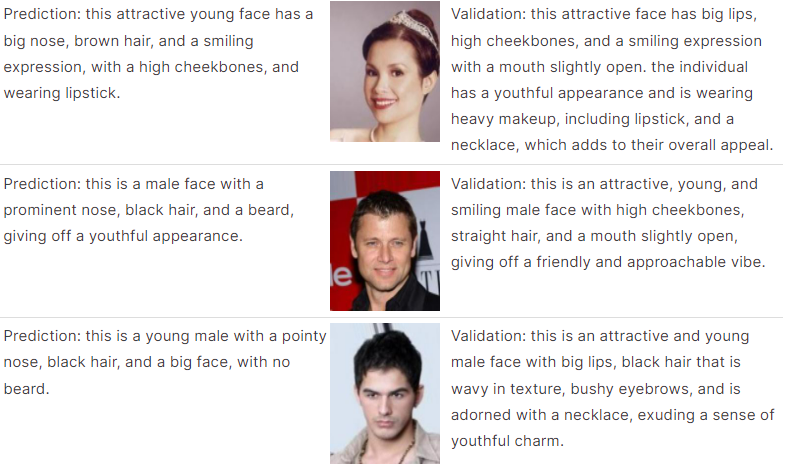}

  \caption{The figure shows a few examples of the predictions produced by FaceGemma and their corresponding validations. The descriptions on the left side of the images are the predictions and on the left are their corresponding ground truth.}

  \label{fig:prediction1}
   \end{center}
 \end{figure}

Figure \ref{fig:prediction1} and \ref{Dataset1} illustrate examples of predictions generated by FaceGemma, showcasing its capability to produce accurate and contextually relevant captions for facial images. These results demonstrate the effectiveness of FaceGemma in leveraging both visual features and facial attributes for enhanced caption generation.

 \begin{figure}[h!]
\begin{center}
  \includegraphics[width=\linewidth]{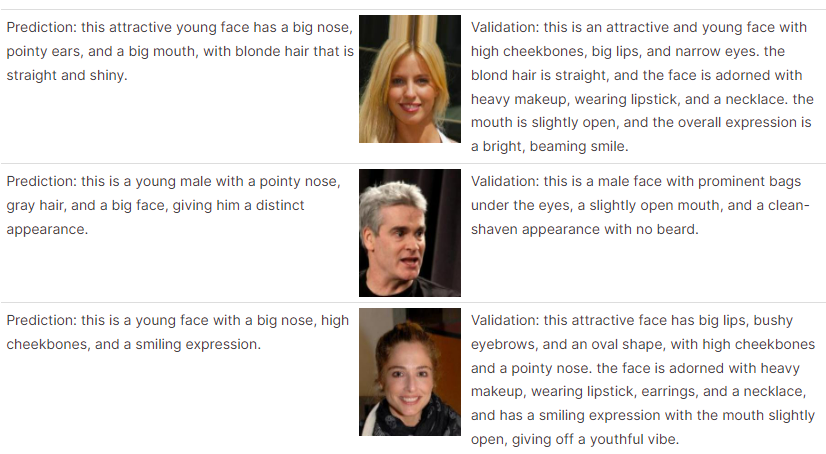}

  \caption{The figure presents several examples of FaceGemma's predictions alongside their corresponding validations. The descriptions on the left side of the images represent the predictions, while those on the right are the actual ground truths.}

  \label{Dataset1}
   \end{center}
 \end{figure}

\section{Conclusion}\label{sec8}

In this research, we introduced the FaceGemma model for generating descriptive and contextually relevant captions for portrait images, focusing on the integration of facial attributes into the captioning process. We conducted extensive experiments, trained our model, and evaluated its performance using BLEU and METEOR scores.

Our results have demonstrated the potential of the FaceGemma model in the field of facial attribute captioning. Additionally, the inclusion of facial attributes significantly improved the model's performance, underlining the importance of considering attribute information in multilingual captioning tasks.

It is worth noting that our research is unique in its focus on captioning based on facial attributes. To the best of our knowledge, there is no prior work in the literature that directly compares with our approach. This highlights the novelty and pioneering nature of our research, breaking new ground in the intersection of computer vision and natural language processing.

Our FaceGemma model, trained and evaluated rigorously, paves the way for diverse applications in the fields of computer vision, image understanding, and multilingual content generation. Its ability to generate contextually relevant captions for portrait images, taking into account both visual features and facial attributes, holds great promise for applications such as image indexing, content retrieval, and accessibility for visually impaired individuals.

As the field of computer vision and natural language processing continues to evolve, our work contributes a valuable step forward in the development of models that can understand and describe the visual world with nuance and accuracy. We anticipate that our research will inspire further exploration in this exciting and groundbreaking area, opening new avenues for innovation and discovery in the realm of image captioning and beyond.


\begin{thebibliography}{9}

\bibitem{imagecaptioning}
Sarang, Poornachandra.
"Image Captioning."
Artificial Neural Networks with TensorFlow 2: ANN Architecture Machine Learning Projects.
Apress, Berkeley, CA, 2021, pp. 471-522.
ISBN: 978-1-4842-6150-7
DOI: 10.1007/978-1-4842-6150-7\_10
URL: https://doi.org/10.1007/978-1-4842-6150-7\_10

\bibitem{computervision}
Srivastava, Rajeev. (2013). \emph{Research Developments in Computer Vision and Image Processing: Methodologies and Applications}. IGI Global.

\bibitem{nlp}
Khurana, Diksha, Aditya Koli, Kiran Khatter, and Sukhdev Singh. 
"Natural Language Processing: State of the Art, Current Trends and Challenges." 
\emph{Multimedia Tools and Applications}, vol. 82, no. 3, pp. 3713--3744, 2023. 
Springer.

\bibitem{dl}
Ian Goodfellow, Yoshua Bengio, and Aaron Courville. 
\emph{Deep Learning}. 
MIT Press, 2016. 
Available online: \url{http://www.deeplearningbook.org}.





\bibitem{en-dec3}
Kelvin Xu, Jimmy Ba, Ryan Kiros, Kyunghyun Cho, Aaron Courville, Ruslan Salakhudinov, Rich Zemel, and Yoshua Bengio.
\emph{Show, attend and tell: Neural image caption generation with visual attention}.
In \emph{International conference on machine learning},
pages 2048--2057, 2015.
Publisher: PMLR.

\bibitem{en-dec4}
Xinyu Xiao, Lingfeng Wang, Kun Ding, Shiming Xiang, and Chunhong Pan.
\emph{Deep Hierarchical Encoder–Decoder Network for Image Captioning}.
\emph{IEEE Transactions on Multimedia}, 
volume 21, number 11, pages 2942-2956, 2019.
DOI: 10.1109/TMM.2019.2915033.

\bibitem{cnn}
Jianxin Wu.
\emph{Introduction to Convolutional Neural Networks}.
\emph{National Key Lab for Novel Software Technology. Nanjing University. China}, 
volume 5, number 23, page 495, 2017.

\bibitem{lstm}
Greg Van Houdt, Carlos Mosquera, and Gonzalo Nápoles.
\emph{A Review on the Long Short-Term Memory Model}.
\emph{Artificial Intelligence Review},
volume 53, pages 5929-5955, 2020.
Publisher: Springer.






\bibitem{deepface}
Omkar M. Parkhi, Andrea Vedaldi, and Andrew Zisserman.
\emph{Deep Face Recognition}.
In \emph{Proceedings of the British Machine Vision Conference (BMVC)},
pages 41.1-41.12, September 2015.
Publisher: BMVA Press.
DOI: 10.5244/C.29.41.
URL: \url{https://dx.doi.org/10.5244/C.29.41}.

\bibitem{sentiattend}
Omid Mohamad Nezami, Mark Dras, Stephen Wan, and Cecile Paris.
\emph{Senti-attend: Image Captioning Using Sentiment and Attention}.
\emph{arXiv preprint arXiv:1811.09789}, 2018.

\bibitem{groupcap}
Fuhai Chen, Rongrong Ji, Xiaoshuai Sun, Yongjian Wu, and Jinsong Su.
\emph{Groupcap: Group-Based Image Captioning with Structured Relevance and Diversity Constraints}.
In \emph{Proceedings of the IEEE conference on computer vision and pattern recognition},
pages 1345-1353, 2018.

\bibitem{facecap}
Omid Mohamad Nezami, Mark Dras, Peter Anderson, and Len Hamey.
\emph{Face-Cap: Image Captioning using Facial Expression Analysis}.
\emph{CoRR},
volume abs/1807.02250, 2018.
URL: \url{http://arxiv.org/abs/1807.02250}.

\bibitem{facetells}
Joanna Hong, Hong Joo Lee, Yelin Kim, and Yong Man Ro.
\emph{Face Tells Detailed Expression: Generating Comprehensive Facial Expression Sentence Through Facial Action Units}.
In \emph{MultiMedia Modeling: 26th International Conference, MMM 2020, Daejeon, South Korea, January 5--8, 2020, Proceedings, Part II},
pages 100-111, 2020.
Publisher: Springer.

\bibitem{faceattend}
Omid Mohamad Nezami, Mark Dras, Stephen Wan, and Cecile Paris.
\emph{Image Captioning Using Facial Expression and Attention}.
\emph{Journal of Artificial Intelligence Research},
volume 68, pages 661-689, 2020.

\bibitem{electronicidentity}
M Usgan, R Ferdiana, and I Ardiyanto.
\emph{Deep Learning Pre-trained Model as Feature Extraction in Facial Recognition for Identification of Electronic Identity Cards by Considering Age Progressing}.
\emph{IOP Conference Series: Materials Science and Engineering},
volume 1115, number 1, pages 012009, 2021.
Publisher: IOP Publishing.
DOI: 10.1088/1757-899X/1115/1/012009.
URL: \url{https://dx.doi.org/10.1088/1757-899X/1115/1/012009}.

\bibitem{transformtell}
Alasdair Tran, Alexander Mathews, and Lexing Xie.
\emph{Transform and Tell: Entity-Aware News Image Captioning}.
In \emph{Proceedings of the IEEE/CVF conference on computer vision and pattern recognition},
pages 13035-13045, 2020.

\bibitem{bornon}
Faisal Muhammad Shah, Mayeesha Humaira, Md Abidur Rahman Khan Jim, Amit Saha Ami, and Shimul Paul.
\emph{Bornon: Bengali Image Captioning with Transformer-based Deep Learning Approach}.
2021.
Eprint: arXiv:2109.05218.
Archive Prefix: arXiv.
Primary Class: cs.CV.

\bibitem{cspdensenet}
Jayakumar Kaliappan, Senthil Kumaran Selvaraj, Baye Molla, and others.
\emph{Caption Generation Based on Emotions Using CSPDenseNet and BiLSTM with Self-Attention}.
\emph{Applied Computational Intelligence \& Soft Computing}, 2022.

\bibitem{explainingemotion}
Oleg Bisikalo, V Kovenko, I Bogach, and O Chorna.
\emph{Explaining Emotional Attitude Through the Task of Image-captioning}.
In \emph{Proceedings of the 6th International Conference on Computational Linguistics and Intelligent Systems (COLINS 2022). Volume I: Main Conference},
Gliwice, Poland, May 12-13, 2022.
Publisher: RWTH Aachen University.

\bibitem{unsupervised}
Zihang Meng, David Yang, Xuefei Cao, Ashish Shah, and Ser-Nam Lim.
\emph{Object-centric Unsupervised Image Captioning}.
In \emph{European Conference on Computer Vision},
pages 219-235, 2022.
Publisher: Springer.

\bibitem{general}
Yinglin Zheng, Hao Yang, Ting Zhang, Jianmin Bao, Dongdong Chen, Yangyu Huang, Lu Yuan, Dong Chen, Ming Zeng, and Fang Wen.
\emph{General Facial Representation Learning in a Visual-Linguistic Manner}.
In \emph{Proceedings of the IEEE/CVF Conference on Computer Vision and Pattern Recognition},
pages 18697-18709, 2022.

\bibitem{attributeclassify}
Sungho Park, Jewook Lee, Pilhyeon Lee, Sunhee Hwang, Dohyung Kim, and Hyeran Byun.
\emph{Fair Contrastive Learning for Facial Attribute Classification}.
In \emph{Proceedings of the IEEE/CVF Conference on Computer Vision and Pattern Recognition},
pages 10389-10398, 2022.

\bibitem{eduvi}
Manisha Gupta, Priya Asthana, and Preetvanti Singh.
\emph{EDUVI: An Educational-Based Visual Question Answering and Image Captioning System for Enhancing the Knowledge of Primary Level Students}.
2023.

\bibitem{faceattributes}
Ziwei Liu, Ping Luo, Xiaogang Wang, and Xiaoou Tang.
\emph{Deep Learning Face Attributes in the Wild}.
\emph{Proceedings of International Conference on Computer Vision (ICCV)},
December 2015.

\bibitem{vggnet}
Karen Simonyan and Andrew Zisserman.
\emph{Very Deep Convolutional Networks for Large-scale Image Recognition}.
\emph{arXiv preprint arXiv:1409.1556},
2014.



\bibitem{resnet50}
Luqman Ali, Fady Alnajjar, Hamad Al Jassmi, Munkhjargal Gocho, Wasif Khan, and M Adel Serhani.
\emph{Performance Evaluation of Deep CNN-Based Crack Detection and Localization Techniques for Concrete Structures}.
\emph{Sensors},
volume 21, number 5, pages 1688, 2021.
Publisher: MDPI.


\bibitem{inceptionv3}
Christian Szegedy, Vincent Vanhoucke, Sergey Ioffe, Jon Shlens, and Zbigniew Wojna.
\emph{Rethinking the Inception Architecture for Computer Vision}.
In \emph{Proceedings of the IEEE conference on computer vision and pattern recognition},
pages 2818-2826, 2016.


\bibitem{gemmateam2024gemmaopenmodelsbased}
Gemma Team, Thomas Mesnard, Cassidy Hardin, et al.
\emph{Gemma: Open Models Based on Gemini Research and Technology}.
Year: 2024.
eprint: 2403.08295.
archivePrefix: arXiv.
primaryClass: cs.CL.
url: \url{https://arxiv.org/abs/2403.08295}

\bibitem{zhai2023sigmoidlosslanguageimage}
Xiaohua Zhai, Basil Mustafa, Alexander Kolesnikov, Lucas Beyer.
\emph{Sigmoid Loss for Language Image Pre-Training}.
Year: 2023.
eprint: 2303.15343.
archivePrefix: arXiv.
primaryClass: cs.CV.
url: \url{https://arxiv.org/abs/2303.15343}

\bibitem{chen2023pali3visionlanguagemodels}
X. Chen et al.
\emph{PaLI-3 Vision Language Models: Smaller, Faster, Stronger}.
Year: 2023.
eprint: 2310.09199.
archivePrefix: arXiv.
url: \url{https://arxiv.org/abs/2310.09199}


\bibitem{bleu}
Kishore Papineni, Salim Roukos, Todd Ward, and Wei-Jing Zhu.
\emph{Bleu: A Method for Automatic Evaluation of Machine Translation}.
In \emph{Proceedings of the 40th annual meeting of the Association for Computational Linguistics},
pages 311-318, 2002.

\bibitem{llama3}
Meta AI. (2024). Meta Llama 3 [Online]. Retrieved July 9, 2024, from [llama3](https://llama.meta.com/)

\bibitem{paligemma}
Google Research. (n.d.). PaliGemma model README. GitHub. Retrieved July 9, 2024, from [PaliGemma](https://github.com/google-research/big\_vision/blob/main/big\_vision/configs/proj/paligemma/README.md)

\bibitem{touvron2023llama}
Touvron, H., Lavril, T., Izacard, G., Martinet, X., Lachaux, M.-A., Lacroix, T., ... \& Joulin, A. (2023). LLaMA: Open and Efficient Foundation Language Models. [arXiv:2302.13971](https://arxiv.org/abs/2302.13971)

\bibitem{meteor}
Satanjeev Banerjee and Alon Lavie.
\emph{METEOR: An Automatic Metric for MT Evaluation with Improved Correlation with Human Judgments}.
In \emph{Proceedings of the acl workshop on intrinsic and extrinsic evaluation measures for machine translation and/or summarization},
pages 65-72, 2005.

\bibitem{faceatt}
Naimul Haque, Iffat Labiba, and Sadia Akter.
``FaceAtt: Enhancing Image Captioning with Facial Attributes for Portrait Images.''
arXiv preprint arXiv:2309.13601 (2023).

\bibitem{faceattdb}
Naimul Haque and Abida Sultana.
``FaceAttDB: A Multilingual Dataset for Facial Attribute Captioning.''
Zenodo, 2023.
doi: 10.5281/zenodo.8144361.





\end{thebibliography}
\end{document}